\documentclass[twocolumn,twocolumn]{IEEEtran}
\usepackage[T1]{fontenc}
\usepackage{color}
\usepackage{mathrsfs}
\usepackage{amsmath}
\usepackage{amsthm} 
\usepackage{amssymb}
\usepackage{graphicx}
\usepackage{cite}
\usepackage{makecell}
\usepackage{algorithm}
\usepackage{algorithmic}
\usepackage{fancyhdr}
\usepackage{diagbox}
\usepackage{bm}
\usepackage[unicode=true,
 bookmarks=true,bookmarksnumbered=true,bookmarksopen=true,bookmarksopenlevel=1,
 breaklinks=false,pdfborder={0 0 0},pdfborderstyle={},backref=false,colorlinks=false]
 {hyperref}
\hypersetup{
 pdftitle={Your Title},
 pdfauthor={Your Name},
 pdfpagelayout=OneColumn, pdfnewwindow=true, pdfstartview=XYZ, plainpages=false}

\usepackage{multirow} 
\usepackage{xcolor}

\allowdisplaybreaks[4]

\usepackage[caption=false,font=footnotesize]{subfig}

\IEEEoverridecommandlockouts

\begin{document}
\bibliographystyle{IEEEtran}

\title{\textcolor{black}{Generative AI-Enhanced Cooperative MEC of UAVs and Ground Stations for Unmanned Surface Vehicles}

\author{\IEEEauthorblockN{Jiahao You$^{\dagger}$, Ziye Jia$^{\dagger \ddagger}$, Chao Dong$^{\dagger}$, Qihui Wu$^{\dagger}$, and Zhu Han$^{\ast \S}$\\}
\IEEEauthorblockA{$^{\dagger}$The Key Laboratory of Dynamic Cognitive System of Electromagnetic Spectrum Space, Ministry of Industry and Information Technology, Nanjing University of Aeronautics and Astronautics, Nanjing, Jiangsu, 211106, China\\
$^{\ddagger}$National Mobile Communications Research Laboratory, Southeast University, Nanjing, Jiangsu, 211111, China\\
$^{\ast}$Department of Electrical and Computer Engineering, University of Houston, Houston, TX 77004, USA\\
$^{\S}$Department of Computer Science and Engineering, Kyung Hee University, Seoul 446-701, South Kore\\
\{yjiahao, jiaziye, dch, wuqihui\}@nuaa.edu.cn, hanzhu22@gmail.com}
}

}

\maketitle
\begin{abstract}
The increasing deployment of unmanned surface vehicles (USVs) require computational support and coverage in applications such as maritime search and rescue. 
Unmanned aerial vehicles (UAVs) can offer low-cost, flexible aerial services, and ground stations (GSs) can provide powerful supports, which can cooperate to help the USVs in complex scenarios.
However, the collaboration between UAVs and GSs for USVs faces challenges of task uncertainties, USVs trajectory uncertainties, heterogeneities, and limited computational resources.
To address these issues, we propose a cooperative UAV and GS based robust multi-access edge computing framework to assist USVs in completing computational tasks. 
Specifically, we formulate the optimization problem of joint task offloading and UAV trajectory to minimize the total execution time, 
which is in the form of mixed integer nonlinear programming and NP-hard to tackle.
Therefore, we propose the algorithm of generative artificial intelligence-enhanced heterogeneous agent proximal policy optimization (GAI-HAPPO). 
The proposed algorithm integrates GAI models to enhance the actor network ability to model complex environments and extract high-level features, thereby allowing the algorithm to predict uncertainties and adapt to dynamic conditions. 
Additionally, GAI stabilizes the critic network, addressing the instability of multi-agent reinforcement learning approaches.
Finally, extensive simulations demonstrate that the proposed algorithm outperforms the existing benchmark methods, thus highlighting the potentials in tackling intricate, cross-domain issues in the considered scenarios.
\end{abstract}


\section{Introduction}\label{s1}
\IEEEPARstart{I}{n} recent years, the increasing demand for unmanned surface vehicles (USVs) in applications such as maritime search and rescue highlights the need for robust computational support and flexible coverage \cite{Survey_iot_2022}. To address these demands, unmanned aerial vehicles (UAVs) offer low-cost, flexible aerial services, while ground stations (GSs) provide powerful computational resources. 
The multi-access edge computing (MEC) framework can provide a distributed computing infrastructure that allows USVs to offload computationally intensive tasks to nearby UAVs and GSs, reducing latency and improving efficiency. Therefore, UAVs and GSs can cooperate within an MEC framework to assist USVs in tackling complex scenarios by complementing each other's capabilities \cite{Cooperative_2024,Luo_DRL_TP_2023, wu_2024_tnse}.

Despite the potential of this collaborative MEC framework, there exist several challenges. USVs often operate under trajectory uncertainties and limited computational resources, while task generation is inherently unpredictable. Additionally, the heterogeneity among USVs, UAVs, and GSs, which arises due to differing communication capabilities, and resource limitations, further complicates effective coordination \cite{Online_2022, NFV_cao_2025, Collaborative_2024}. These challenges necessitate advanced MEC frameworks capable of handling task uncertainties, dynamic conditions, and heterogeneous environments.

Existing cooperative schemes for USVs, UAVs, and GSs within the MEC paradigm primarily focus on task allocation \cite{Integrating_2024} or hybrid communication networks \cite{3u_2023}. While these approaches improve coordination and enable computational offloading, they often struggle to address the inherent uncertainty and complexity of real-world applications. Multi-agent reinforcement learning (MARL) emerges as a promising approach within MEC, enabling agents to adaptively interact with dynamic environments \cite{Beyond_2024,VR_MARL_2024}. However, traditional MARL algorithms face limitations in extracting high-level features and maintaining training stability in the presence of environmental uncertainties.
Generative artificial intelligence (GAI) offers a compelling solution to these challenges. By leveraging GAI models such as generative adversarial network (GAN) and transformer, MARL algorithms in MEC scenarios can achieve higher sampling efficiency, better generalization capabilities, and enhanced robustness to dynamic conditions. Specifically, GAI improves feature extractions in actor networks and stabilizes training for critic networks, addressing the limitations of traditional MARL \cite{GAI_2024, Approximation_2022}.

In this work, we propose a cooperative UAV and GS-based robust MEC framework to support USVs in completing computational tasks. The framework leverages the flexibility of UAVs and the computational power of GSs within an MEC environment to handle task uncertainties and USV trajectory randomness effectively. To optimize the collaboration, we formulate a mixed-integer nonlinear programming (MINLP) problem for joint task offloading and UAV trajectory planning, aiming to minimize total execution time. To effectively solve the problem, we design a GAI-enhanced heterogeneous agent proximal policy optimization (GAI-HAPPO) algorithm. By integrating GAI, the algorithm improves the actor network ability to model complex environments and predict uncertainties while stabilizing the critic network. Extensive simulations demonstrate that GAI-HAPPO outperforms existing benchmark methods, achieving a 22.8\% performance improvement and effectively tackling intricate cross-domain challenges.

The rest of the paper is organized as follows.
Section \ref{s2} details the system model and problem formulation. 
Section \ref{s3} introduces the GAI-HAPPO algorithm. 
Section \ref{s4} presents numerical results, and conclusions are drawn in Section \ref{s5}.

\section{System Model}\label{s2}
\begin{figure}[!t]
\centering
\includegraphics[width=8cm]{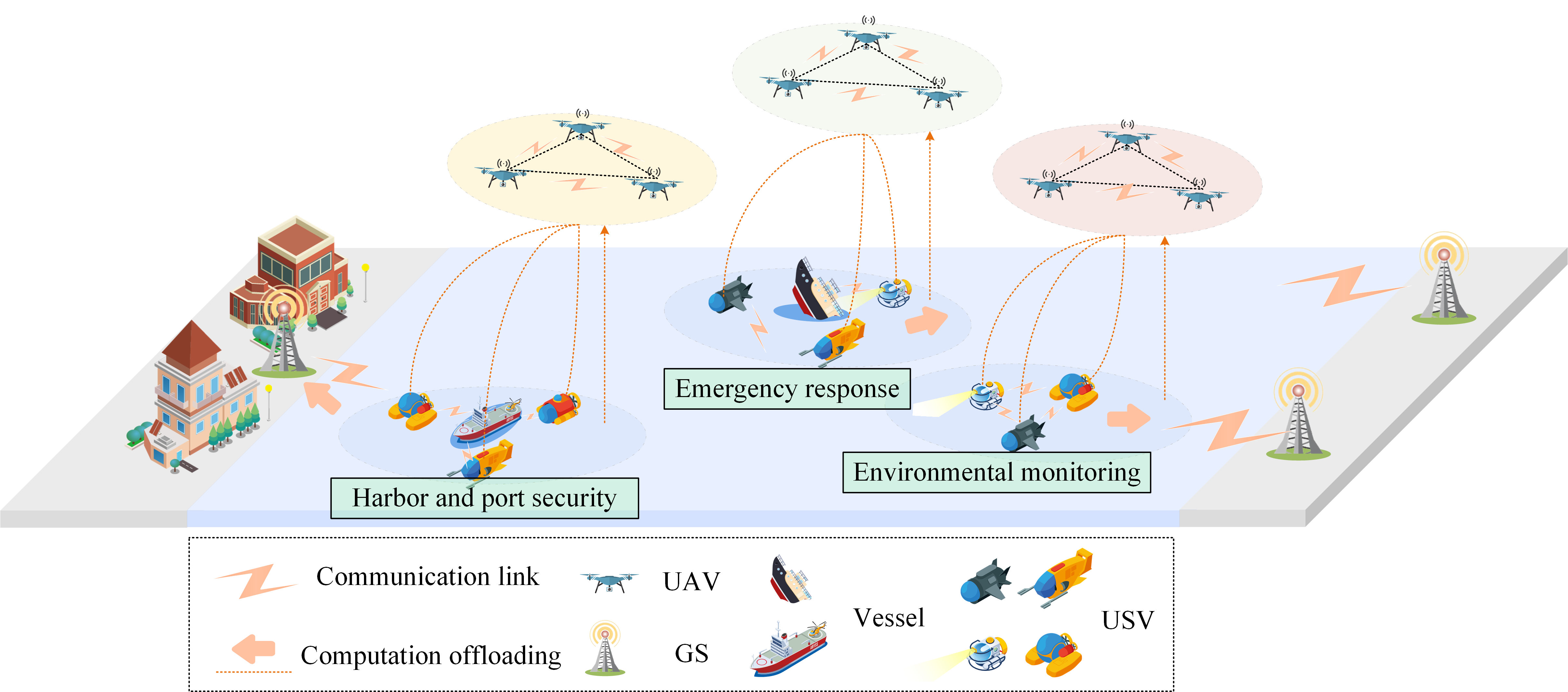}
\caption{System model overview.}
\label{f1}
\end{figure}

As shown in Fig. \ref{f1}, the MEC framework consists of USVs, UAVs, and GSs. UAVs are equipped with MEC servers to assist USVs, while GSs provide powerful computational capabilities. The sets of USVs, UAVs, and GSs are denoted as $\mathbb{S} = \{1, 2, ..., I\}$, $\mathbb{U} = \{1, 2, ..., J\}$, and $\mathbb{G} = \{1, 2, ..., K\}$, respectively. The system time duration is divided into discrete time slots $\mathbb{T} = \{1, 2, ..., T\}$ with each slot of duration $\tau$. Each USV $i \in \mathbb{S}$ generates a task $A_i(t) = \{d_i(t), c_i(t)\}$ at time $t$, where $d_i(t)$ is the data size, and $c_i(t)$ is the required computation in CPU cycles per bit. Task generation follows a poisson distribution with mean $\lambda_i$.
USVs perform tasks such as environmental monitoring and emergency response. The task execution involves three stages: local computation on USVs, transmission and computation on UAVs, and transmission and computation on GSs. 

\subsection{Communication Model}
\subsubsection{USV to UAV (U2U)}
The path loss between USV $i$ and UAV $j$ is modeled as
\begin{equation}
\begin{split}
\xi_{i,j}(t) &= \frac{\zeta_{L}-\zeta_{NL}}{1+a \exp \{-b [\varsigma_{i,j}(t) -a \}} \\
& + 20 \lg\left(\frac{4\pi f_c\|l_i^s(t)-l_j^u(t)\| }{C}\right)+\zeta_{NL},
\end{split}
\end{equation}
in which $f_c$ denotes the carrier frequency, and $C$ represents the speed of light.
$\zeta_{L}$, $\zeta_{NL}$, $a$, and $b$ are parameters characterizing the environment \cite{Jia_Joint_HAP_2021}. The average rate between USV $i$ and UAV $j$ is computed as 
\begin{equation}
R_{i,j}^{u2u}(t) = B_{i,j}^{u2u}(t) \log_2 \left(1 + \frac{P_i^s \xi_{i,j}(t)}{N_G}\right).
\end{equation}
where $B_{i,j}^{u2u}(t)$ is the available bandwidth at UAV $j$ from USV $i$, $P_{i}^s$ represents the transmitting power of USV $i$, and $N_G$ indicates the power of the additive white Gaussian noise.

\subsubsection{USV to GS (U2G)}
The channel power gain from USV $i$ to GS $k$ is
\begin{equation}
G_{i,k}(t) = G_0 (\|l_i^s(t) - l_k^g(t)\|)^{-2},
\end{equation}
where $G_0$ is the channel power gain when $\|l_i^s(t)-l_k^g(t)\|_2$ equals 1m \cite{Luo_DRL_TP_2023}. 
Consequently, the transmission rate from USV $i$ to GS $k$ is 
\begin{equation}
R_{i,k}^{u2g}(t) = B_{i,k}^{u2g} \log_2 \left(1 + \frac{P_i^s G_{i,k}(t)}{N_G}\right).
\end{equation}
where $B_{i,k}^{u2g}(t)$ represents the available bandwidth at GS $k$ from USV $i$.

\subsection{USV Mobility Model}
The positions of USVs, UAVs, and GSs are depicted by a three-dimensional cartesian coordinate system, i.e., $l_i^s(t) = (x_i^{s}(t), y_i^{s}(t), h_i^s(t))$, $l^u_j(t) = (x_j^{u}(t), y_j^{u}(t),h_j^u(t))$, and $l^g_k(t) = (x_k^{g}(t), y_k^{g}(t), h_k^g(t))$, respectively.
It is assumed that USVs follow the Gauss-Markov mobility model \cite{Mobility_2015,Online_2022}.
Specifically, the velocity of USV $i$ at time slot $t$ can be described as
\begin{equation}
v_i(t+1) = \mu  v_i(t) + (1 - \mu) \overline{v} + \overline{\sigma} \sqrt{1-\mu^2}w_i(t),
\end{equation}
where $v_i(t) = (v_i^x(t),v_i^y(t), 0)$ denotes the velocity vector and $w_i(t)=(w_i^x(t),w_i^y(t), 0)$ represents an uncorrelated random Gaussian process $\mathcal{N}(0,\sigma^2)$. $\mu$, $\overline{v}$, and $\overline{\sigma}$ denote the memory level, asymptotic mean, and asymptotic standard deviation of velocity, respectively. Therefore, the location is updated as
\begin{equation}
l^s_i(t+1) = l^s_i(t) + v_i(t)\tau.
\end{equation}

\subsection{UAV Mobility Model}
To save energy, UAVs maintain a constant altitude $H$ above the ground, thereby preventing frequent ascents and descents.
Consequently, we consider the horizontal flight of the UAV over the target area, which requires modeling the UAV's flight dynamics.
In time slot $t$, the flying horizontal azimuth $\theta_j(t)$ of UAV $j$ and the flying distance $k_j(t)$ should satisfy the following constraints
\begin{equation}
\theta_j(t) \in [0, 2\pi], \forall j \in \mathbb{U}, \forall t \in \mathbb{T},\label{e6}
\end{equation}
and
\begin{equation}
k_j(t) \in [0, k^{max}], \forall j \in \mathbb{U}, \forall t \in \mathbb{T},\label{e7}
\end{equation}
where $k^{max}$ represents the maximum distance a UAV can fly in time slot $t$, and the flying distance is calculated as
\begin{equation}
k_j(t) = \| l^u_j(t+1) - l^u_j(t)\|.
\end{equation}

Additionally, the horizontal location coordinates $x$ and $y$ of UAV $j$ at time $t$ can be determined using the horizontal azimuth $\theta_j(t)$ and the flying distance $k_j(t)$ as 
\begin{equation}
x_j^{u}(t) = x_j^{u}(0) + \sum_{t=1}^{t} k_j(t) \cos(\theta_j(t)),
\end{equation}
and 
\begin{equation}
y_j^{u}(t) = y_j^{u}(0) + \sum_{t=1}^{t} k_j(t) \sin(\theta_j(t)).
\end{equation}

To ensure UAVs stay within the target area, which affect trajectory decisions, the following constraints should be satisfied: 
\begin{equation}
0 \leq x_j^{u}(t) \leq x_{max}, \forall j \in \mathbb{U}, t \in \mathbb{T},\label{e13}
\end{equation}
and
\begin{equation}
0 \leq y_j^{u}(t) \leq y_{max}, \forall j \in \mathbb{U}, t \in \mathbb{T},\label{e14}
\end{equation}
where $x_{max}$ and $y_{max}$ denote the maximum X and Y coordinate values of the target area, respectively.

\subsection{Computation Model}
In each time slot $t$, the task $A_i(t)$ generated by USV $i$ can be processed locally or offloaded to a UAV $j$ or a GS $k$. The offloading decisions are represented by binary variables
\begin{equation}
p_{i,j}(t) = 
\begin{cases}
1, & \text{if USV } i \text{ offloads to UAV } j, \\
0, & \text{otherwise.}
\end{cases}
\end{equation}
and
\begin{equation}
q_{i,k}(t) = 
\begin{cases}
1, & \text{if USV } i \text{ offloads to GS } k, \\
0, & \text{otherwise.}
\end{cases}
\end{equation}

It is worth noting that for task $A_i(t)$, only one UAV and one GS can be selected for offloading computation within the same time slot, i.e.,
\begin{equation}
\sum_{j=1}^J p_{i,j}(t) \leq 1,\forall i \in \mathbb{S}, t\in\mathbb{T},\label{e17}
\end{equation}
and
\begin{equation}
\sum_{k=1}^K q_{i,k}(t) \leq 1, \forall i \in \mathbb{S},t\in\mathbb{T}.\label{e18}
\end{equation}

Let $\alpha_i(t)$, $\beta_i(t)$, and $\gamma_i(t)$ denote the proportions of workload $A_i(t)$ processed by USV $i$, UAV $j$, and GS $k$, respectively, such that $\alpha_i(t) + \beta_i(t) + \gamma_i(t) = 1$. The total workload can be expressed as
\begin{equation}
A_i(t) = \alpha_i(t) A_i(t) + \beta_i(t) A_i(t) + \gamma_i(t) A_i(t).
\end{equation}

At each time slot $t$, the dynamic changes in task queues for USVs, UAVs, and GSs are updated as
\begin{equation}
Q_i^s(t+1) = \max \{0, Q_i^s(t) + \alpha_i(t) d_i(t) - \tau f_i^s\},
\end{equation}
\begin{equation}
Q_j^u(t+1) = \max\{0, \sum_{i=1}^{I} p_{i,j}(t) Q_j^u(t) + \beta_{i}(t)d_i(t) - \tau f_j^u\},
\end{equation}
and
\begin{equation}
Q_k^g(t+1) = \max\{0, \sum_{i=1}^{I} q_{i,k}(t) Q_k^g(t) + \gamma_{i}(t)d_i(t) - \tau f_k^g\},
\end{equation}
where $f_i^s$, $f_j^u$, and $f_k^g$ denote the computation resources of USV $i$, UAV $j$, and GS $k$, respectively.

\subsubsection{USV based Computing}
The local computation time for USV $i$ is
\begin{equation}
T_i^s(t) = \frac{Q_i^s(t)c_i(t)}{f_i^s} + \frac{\alpha_i(t)d_i(t)c_i(t)}{f_i^s}.
\end{equation}

\subsubsection{UAV based Computing}
If a task is offloaded to UAV $j$, the computation delay is
\begin{equation}
\begin{aligned}
T_{i,j}^{u}(t) = p_{i,j}(t) \bigg(\frac{Q_j^u(t)c_i(t)}{f_j^u}&+\frac{\beta_{i}(t)d_i(t)}{R_{i,j}^{u2u}(t)}
\\ &\qquad+\frac{\beta_{i}(t) d_i(t) c_i(t)}{f_j^u}\bigg).
\end{aligned}
\end{equation}

\subsubsection{GS based Computing}
For tasks offloaded to GS $k$, the computation delay is
\begin{equation}
\begin{aligned}
T_{i,k}^{g}(t) = q_{i,k}(t) \bigg(\frac{Q_k^g(t)c_i(t)}{f_k^g}&+\frac{\gamma_{i}(t)d_i(t)}{R_{i,k}^{u2g}} 
\\ &\qquad+\frac{\gamma_{i}(t) d_i(t) c_i(t)}{f_k^g}\bigg),
\end{aligned}
\end{equation}

As such, tasks can be processed on USVs, UAVs and GSs.	
Hence, the delay for the tasks generated by USV $i$ in time slot $t$ is calculated as
\begin{equation}
T_{i}^{t}(t) = T_{i}^{s}(t) + \sum_{j=1}^J T_{i,j}^{u}(t) + \sum_{k=1}^K T_{i,k}^{g}(t).
\end{equation}

Consequently, the cumulative time cost for all USV-generated tasks in $T$ is 
\begin{equation}
\Phi = \sum_{t=1}^{T}\sum_{i=1}^I T_{i}^{t}(t).
\end{equation}

\subsection{Problem Formulation}
The optimization problem is formulated is to minimize the total execution time by jointly optimizing the flight trajectories of UAVs (i.e., $\mathbb{L} = \{\theta_j(t), k_j(t)| \forall j \in \mathbb{U}, t \in \mathbb{T}\}$), offloading decisions (i.e., $\mathbb{P} = \{p_{i,j}(t)| \forall i \in \mathbb{S}, j \in \mathbb{U}, t \in \mathbb{T}\} $ and $\mathbb{Q} = \{q_{i,k}| \forall i \in \mathbb{S}, k \in \mathbb{G}, t \in \mathbb{T}\}$), and the task execution ratios (i.e., $\pmb{\alpha} = \{\alpha_i(t)| \forall i \in \mathbb{S}, t \in \mathbb{T}\}, \pmb{\beta} = \{\beta_i(t)| \forall i \in \mathbb{S}, t \in \mathbb{T}\},$ and $ \pmb{\gamma} = \{\gamma_i(t)| \forall i \in \mathbb{S}, t \in \mathbb{T}\}$), i.e., 
\begin{subequations}
\begin{align}
\mathscr{P}0: &\min_{\mathbb{L},\mathbb{P},\mathbb{Q},\pmb{\alpha},\pmb{\beta},\pmb{\gamma}}\sum_{t=1}^{T}\sum_{i=1}^I T_{i}^{t}(t) \notag\\
&\text{s.t. } (\ref{e6}),(\ref{e7}),(\ref{e13})-(\ref{e18}), \notag\\
\end{align}
\end{subequations}
Note that $\mathscr{P}0$ is a MINLP, as it includes a non-convex objective function and discrete variables. Additionally, the dynamic positions of USVs and UAVs, random task arrivals, and uncertain resources available in each time slot increase the dynamic characteristics, which poses significant challenges for traditional optimization algorithms.

\section{Algorithm Design}\label{s3}
To address the aforementioned issues and achieve efficient task offloading and UAV trajectory optimization, we design the GAI-HAPPO algorithm.

\subsection{MG Formulation} 
The problem can be modeled as a Markov game.
USVs and UAVs can be regarded as $N=I+J$ agents, thus the problem is represented by the tuple $\langle \mathcal{S}, \{\mathcal{O}_i^s, \mathcal{O}_j^u\}_{i \in \mathbb{S}, j \in \mathbb{U}}, \{\mathcal{A}_i^s, \mathcal{A}_j^u\}_{i \in \mathbb{S}, j \in \mathbb{U}}, \mathcal{R}, \gamma \rangle$.

\subsubsection{State Space}
The state space represents the comprehensive set of variables that define the current status of the system at time $t$, i.e.,
\begin{equation}
\begin{aligned}
s(t) = \{l_i^s(t), &l_j^u(t), l_k^g(t),\\
&Q_i^s(t),Q_j^u(t),Q_k^g(t)|i \in \mathbb{S}, j \in \mathbb{U}, k \in \mathbb{G}\}.
\end{aligned}
\end{equation}

\subsubsection{Observation Space}
Each agent (USV or UAV) has its own observation space.

\paragraph{USV agent}
The USV agent focuses on local computation and task offloading. Its observation space is
\begin{equation}
\begin{aligned}
o_i^s(t) = \{l_i^s(t), l_j^u(t), &l_k^g(t), Q_i^s(t),Q_j^u(t),\\
&Q_k^g(t),A_i(t) | i \in \mathbb{S}, j \in \mathbb{U}, k \in \mathbb{G}\}.
\end{aligned}
\end{equation}

\paragraph{UAV agent}
The UAV agent observation space is
\begin{equation}
\begin{aligned}
o_j^u(t) = \{l_i^s(t), l_j^u(t), &l_k^g(t), Q_i^s(t),Q_j^u(t),\\
&Q_k^g(t),A_i(t) | i \in \mathbb{S}, j \in \mathbb{U}, k \in \mathbb{G}\}.
\end{aligned}
\end{equation}

\subsubsection{Action Space}
The action space defines the set of possible actions an agent can take at time $t$.
\paragraph{USV agent}
The action space of the USV agent is represented as
\begin{equation}
\begin{aligned}
a_i^s(t) = \{p_{i,j}(t), q_{i,k}(t), &\alpha_i(t), \\
&\beta_i(t), \gamma_i(t) | i \in \mathbb{S}, j \in \mathbb{U}, k \in \mathbb{G} \}.
\end{aligned}
\end{equation}
where the UAV needs to choose which UAV and GS to offload the tasks, as well as how to allocate the task proportions. 

\paragraph{UAV agent}
The action space of the UAV agent is
\begin{equation}
a_j^u(t) = \{\theta_j(t), k_j(t)\}.
\end{equation}
where the UAV needs to determine the flying azimuth and distance.

\subsubsection{Reward Function}
The reward function is based on the time required to process a single bit, considering the uncertainty of tasks and the complex variability of the environment. The reward function $r(t)$ can be expressed as
\begin{equation}
r(t) = \sum_{i=1}^{I}d_i(t)/T_i^t(t).
\end{equation}

\begin{algorithm}[!t]
\caption{GAI-HAPPO}
\label{algorithmic1}
\begin{algorithmic}[1]
\REQUIRE Batch size $B$, number of agents $n$, episodes $K$, and steps per episode $T$.
\STATE \textbf{Initialize:} Actor networks $\theta^{i_m}$, generator network $G$, and discriminator network $D$. \label{a1}
\FOR{$k = 0$ to $K-1$}
    \STATE Collect trajectories by running the joint policy $\boldsymbol{\pi}_{\boldsymbol{\theta}_k}=(\pi_{\theta_k^1}^1, \ldots, \pi_{\theta_k^n}^n)$. \label{a3}
    \STATE Store transitions $\{(s_t, o_t^i, a_t^i, r_t, s_{t+1}, o_{t+1}^i),$ $\forall i \in N, t \in T\}$ in $B$. \label{a4}
    \STATE Calculate the advantage function $\hat{A}(\mathrm{s}, \mathbf{a})$ using generalized advantage estimation.\label{a5}
    \STATE Generate a random permutation of agents $i_{1: n}$.\label{a6}
    \STATE Set $M^{i_1}(\mathrm{s}, \mathbf{a})=\hat{A}(\mathrm{s}, \mathbf{a})$.\label{a7}
    \FOR{agent $i_m = i_1, \ldots, i_n$}
        \STATE Update actor $i_m$ with $\theta_{k+1}^{i_m}$, the argmax of the clip objective.\label{a9}
        \STATE Calculate $M^{i_{1: m+1}}(s, \bm{a})$. \label{a10}
    \ENDFOR
    \STATE Update generator network $G$ and discriminator network $D$. \label{a12}
\ENDFOR
\end{algorithmic}
\end{algorithm}

\subsection{GAI-HAPPO}
GAI-HAPPO integrates transformer-based actor networks and GAN-based critic networks to address heterogeneous multi-agent collaboration problems in MEC scenarios.
Using linear transformations, the input data is transformed into query, key, and value matrices for self-attention calculations.
\begin{equation}
\operatorname{Attention}(Q, K, V) = \operatorname{softmax}\left(\frac{Q K^T}{\sqrt{d_k}}\right)V,
\end{equation}
allowing the model to capture complex dependencies across states. The output is refined through fully connected layers to predict task ratios, associations, and trajectory parameters.

For the critic network, a GAN framework is employed to enhance stability and action-value estimation. The generator predicts state values \(\hat{V}(s_t)\), while the discriminator refines predictions through adversarial training. Their respective loss functions are defined as 
\begin{equation}
L_D = -\mathbb{E}\left[\log D(s_t, V^*(s_t)) + \log (1 - D(s_t, \hat{V}(s_t)))\right],
\end{equation}
and
\begin{equation}
L_G = \mathbb{E}\left[\log(1 - D(s_t, \hat{V}(s_t)))\right],
\end{equation}
ensuring improved accuracy in value estimation.

\begin{figure}[!t]
\centering
\includegraphics[width=7cm]{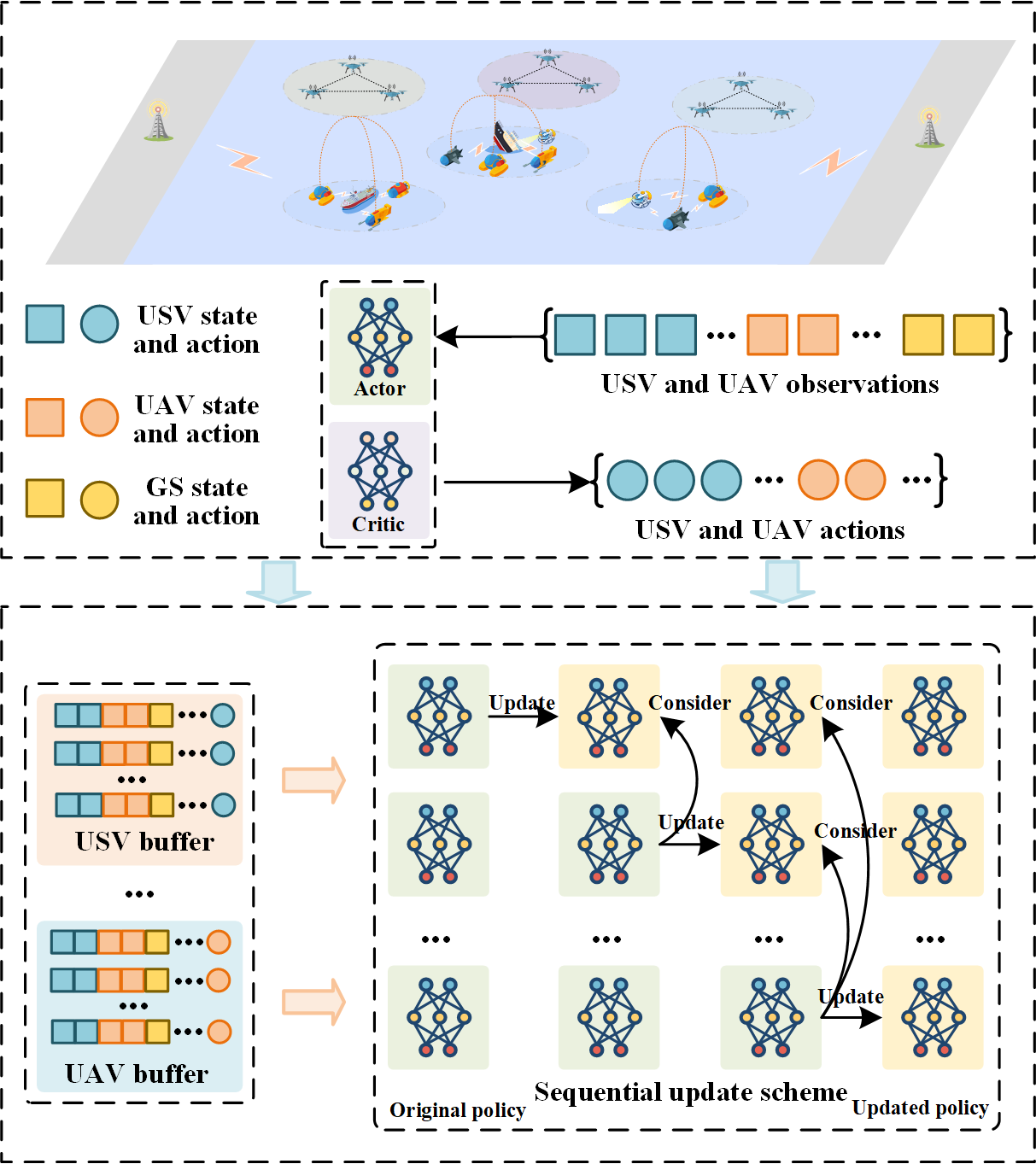}
\caption{The GAI-HAPPO framework in cooperative UAV-GS MEC.}
\label{f2}
\end{figure}

Algorithm \ref{algorithmic1} leverages GAI to enhance MARL via a coordinated policy optimization process. 
As illustrated in Fig. \ref{f2}, GAI-HAPPO uses a sequential update scheme to improve the collaboration in heterogeneous multi-agent systems \cite{HARL_2024}.	
Each agent policy $\bm{\pi}^i_{\theta^i}$ is parameterized by $\theta^i$, creating a joint policy $\bm{\pi}$ parameterized by $\bm{\theta} = \{\theta^1,...,\theta^n\}$.
During each iteration $k + 1$, with a permutation of agents $i_{1: n}$, each agent $i_{m}$ (where $m \in \{1,...,n\}$) sequentially optimizes its policy parameter.	
This optimization involves agent $i_m$ selecting a policy parameter $\theta_{k+1}^{i_m}$ that maximizes the following objective, i.e.,
\begin{equation}
\begin{split}
&\mathcal{L}(\theta_{k+1}^{i_m}) = \mathbb{E}_{s \sim \rho_{\bm{\pi}_{\bm{\theta}^k}}, \bm{a}\sim \bm{\pi}_{\bm{\theta}_{k}}}   \bigg[ \min \bigg( \frac{\pi_{i_m}^{\theta_{i_m}}(a_{i_m}|s)}{\pi_{i_m}^{\theta_{i_m}^k}(a_{i_m}|s)} M^{i_{i:m}}(s, \bm{a}), \\
&\qquad\qquad\qquad \text{clip}\bigg( \frac{\pi_{i_m}^{\theta_{i_m}}(a_{i_m}|s)}{\pi_{i_m}^{\theta_{i_m}^k}(a_{i_m}|s)}, 1 \pm \epsilon \bigg)M^{i_{i:m}}(s, \bm{a})\bigg)\bigg],
\end{split}\label{eq33}
\end{equation}
where
\begin{equation}
M^{i_{1: m}}(s, \bm{a})=\frac{\boldsymbol{\pi}_{\boldsymbol{\theta}_{k+1}^{i_{1: m-1}}}^{i_{1: m-1}}\left(\bm{a}^{i_{1: m-1}} \mid s\right)}{\boldsymbol{\pi}_{\boldsymbol{\theta}_k^{i_{1: m}}}^{i_{1:m-1}}\left(\bm{a}^{i_{1: m-1}} \mid s\right)} \hat{A}(s, \bm{a}).
\label{eq34}
\end{equation}
$\hat{A}(s, \bm{a})$ is the advantage function. 
$M^{i_{1 m}}$ represents the advantage function in GAI-HAPPO, which needs to consider the parameters of the other agents. 
$\epsilon$ denotes a hyperparameter used to control the magnitude of the policy updates.

The generator network is updated to minimize the following loss function $\theta_G$, effectively replacing the V-value network update, i.e.,
\begin{equation}
\theta_G = \arg \min \frac{1}{B T} \sum_{b=1}^B \sum_{t=0}^T\left(G\left(s_t\right)-\hat{R}_t\right)^2.
\end{equation}

\begin{figure}[!t]
\centering
\includegraphics[width=7cm]{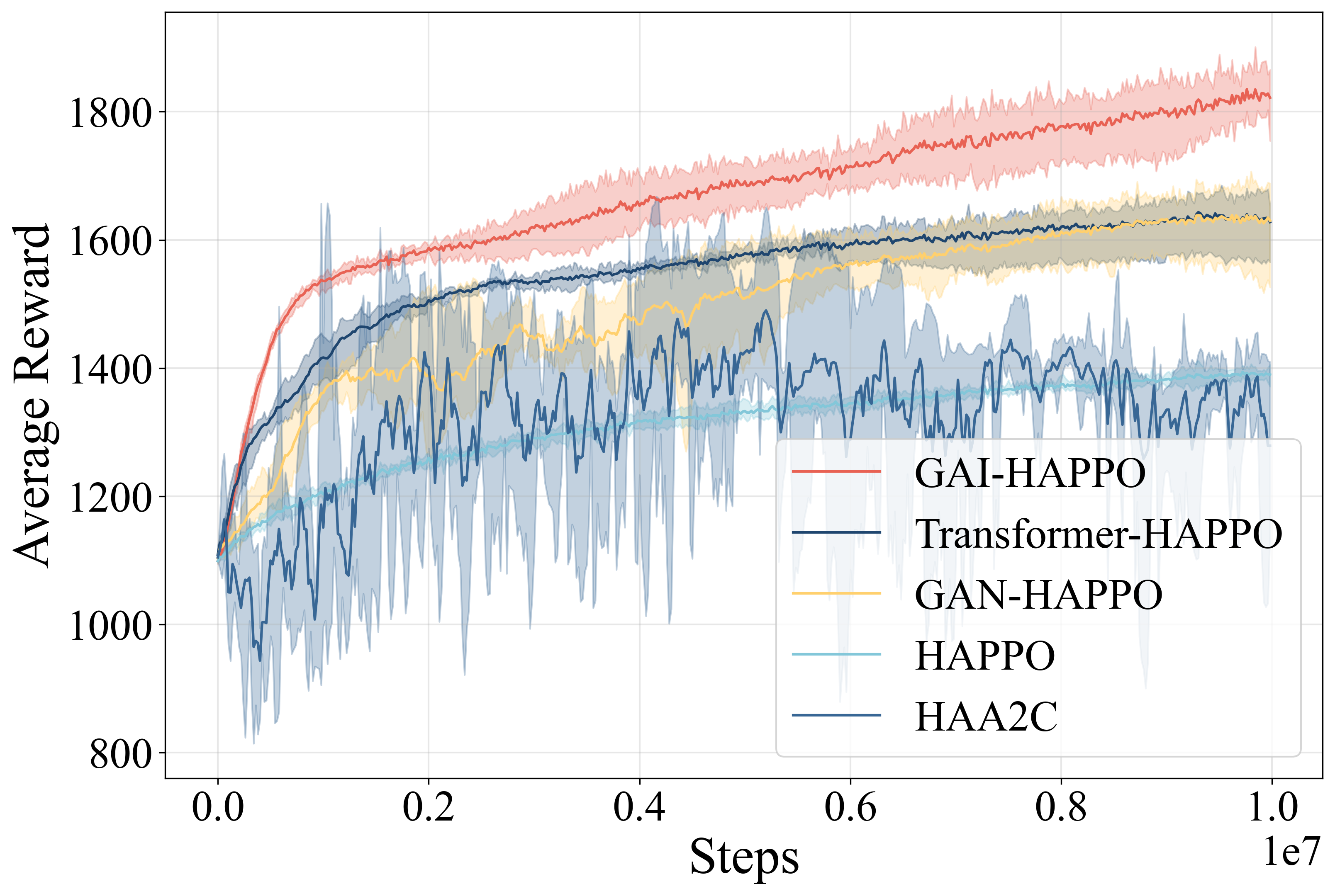}
\caption{Average reward of standard setting.}
\label{f3}
\end{figure}

\section{Simulation Results}\label{s4}
To validate the effectiveness of the proposed GAI-HAPPO algorithm, we conduct numerical experiments and compare its performance with four benchmark algorithms:

\begin{itemize}
    \item \textbf{HAA2C}: Expands the advantage actor-critic framework for heterogeneous MARL agents \cite{HARL_2024}.
    \item \textbf{HAPPO}: An evolution of PPO, designed for collaboration in heterogeneous MARL systems \cite{HARL_2024}.
    \item \textbf{GAN-HAPPO}: Combines GAN with HAPPO to enhance the stability and accuracy of the critic network.
    \item \textbf{Transformer-HAPPO}: Introduces transformer architectures into HAPPO to address sequential and spatial-temporal challenges in MARL.
\end{itemize}

\subsection{Parameter Setting}
We simulate a \(1,000m \times 1,000m\) area with 6 USVs generating tasks, supported by 4 UAVs and 2 GSs for task processing. Each USV generates a computation-intensive task per time slot, and UAVs adjust positions dynamically to optimize task execution. 
Key parameters include path loss parameters $\zeta_L = 2.3$, $\zeta_{NL} = 34$, and the mean task arrival rate $\lambda_i = 15$. The transmitting power is set to $P_i^s = 1$ W, and the maximum UAV flying distance per time slot is $k^{max} = 30$ m. The computation complexity for each task is $c_i(t) = 270$ cycles/bit. The noise power is $N_G = -114$ dBm. The learning rates for the algorithms are set to $5 \times 10^{-5}$ and $10^{-4}$, respectively.

\subsection{Evaluation on the Training of GAI-HAPPO and Benchmark Algorithms}
Fig. \ref{f3} shows the training results for GAI-HAPPO and benchmark algorithms. GAI-HAPPO achieves the highest average rewards with excellent stability and convergence after 2,500 episodes. Transformer-HAPPO follows closely but exhibits greater volatility. GAN-HAPPO improves upon HAPPO, while HAA2C shows the weakest performance with lower rewards and poor convergence. These results highlight GAI-HAPPO's advantage in handling heterogeneous environments and dynamic conditions.

\begin{figure*}[htbp]
\centering
\subfloat[Average execution delay.]{
\includegraphics[width=4.3cm]{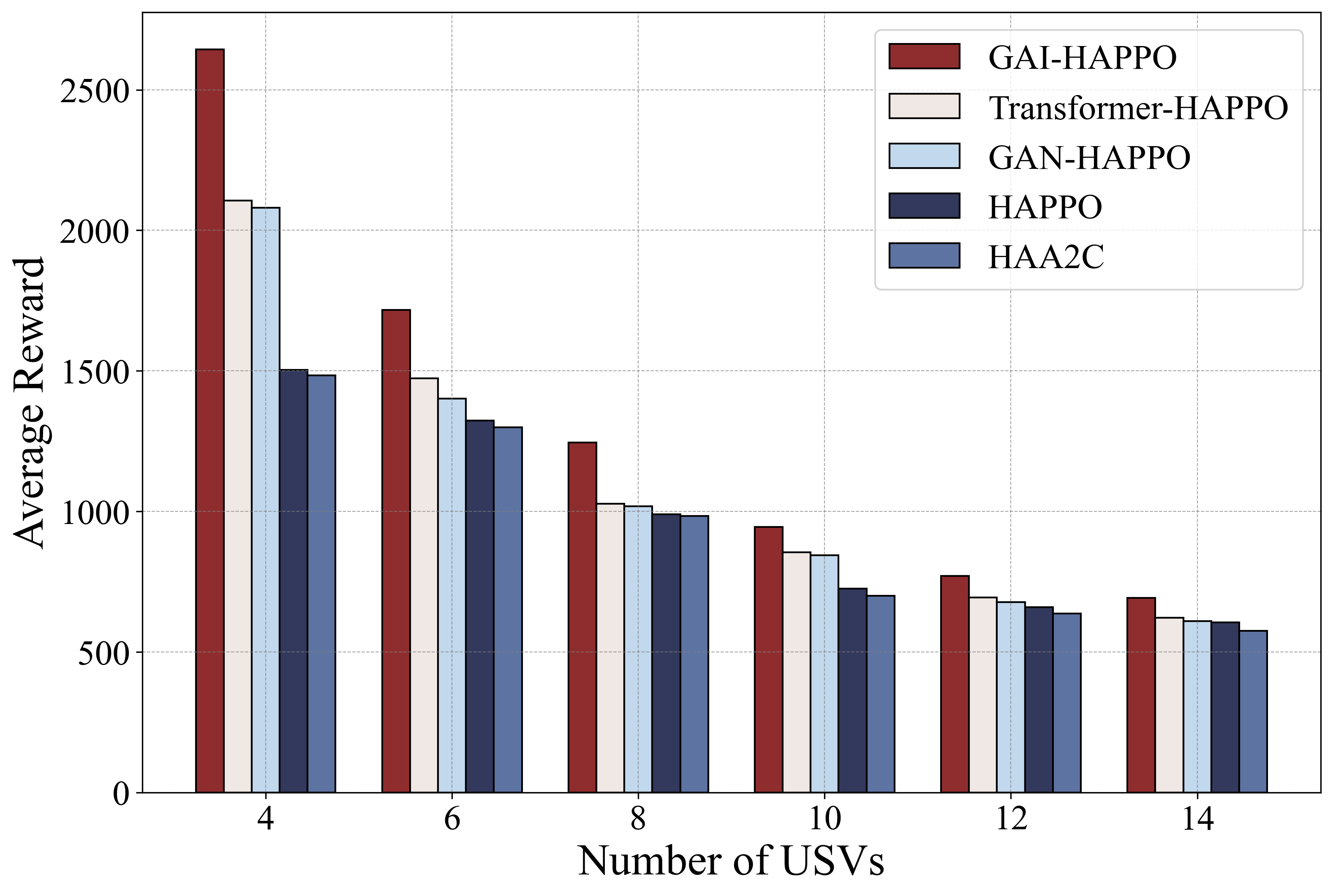}\label{f4a}
}
\hspace{-0.3cm}
\subfloat[Average reward.]{
\includegraphics[width=4.3cm]{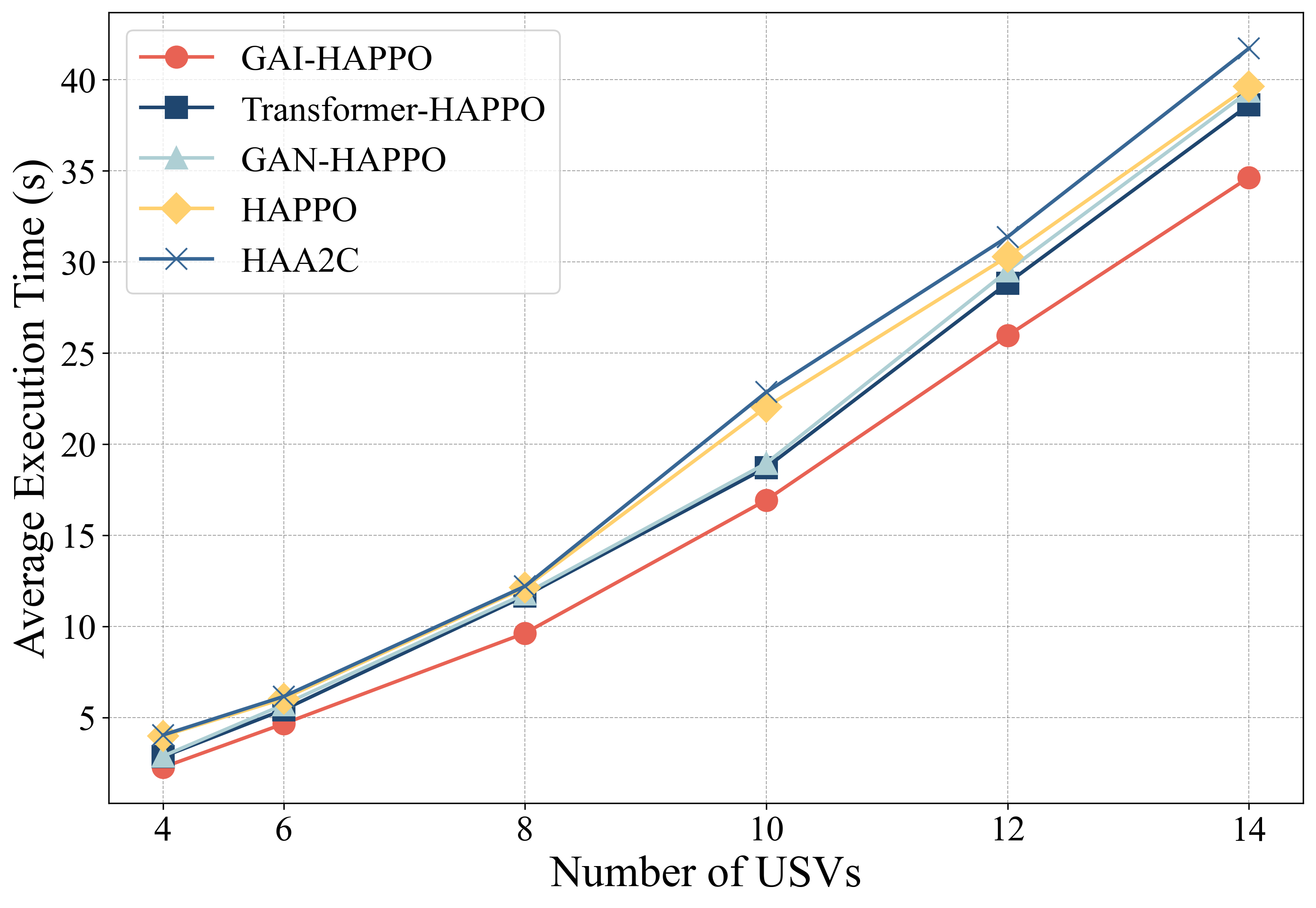}\label{f4b}
}
\hspace{-0.3cm}
\subfloat[Average execution delay.]{
\includegraphics[width=4.3cm]{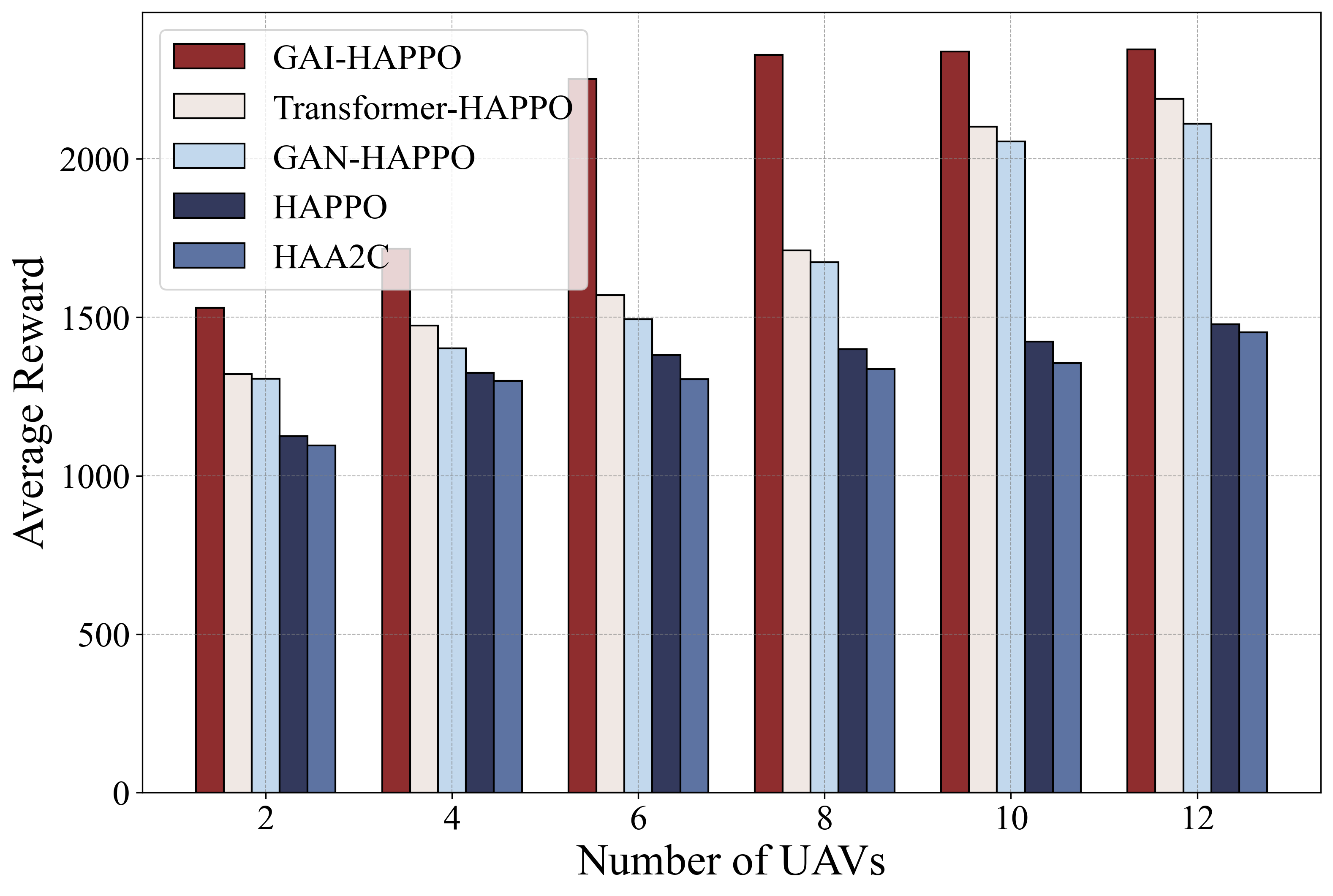}\label{f4c}
}
\hspace{-0.3cm}
\subfloat[Average reward.]{
\includegraphics[width=4.3cm,height=3.1cm]{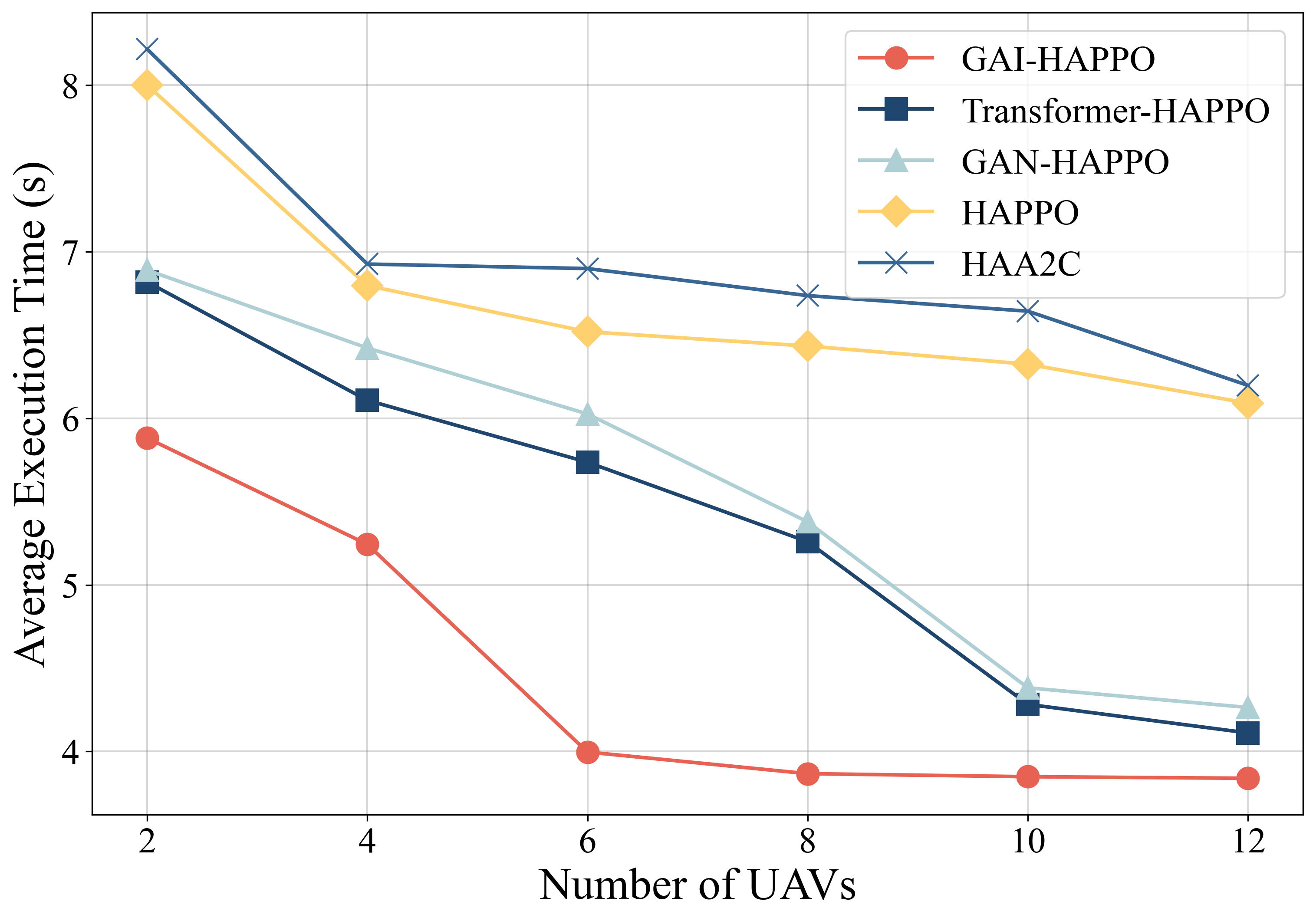}\label{f4d}
}
\caption{Performance with different numbers of USVs and UAVs.}
\label{f4}
\end{figure*}

\subsection{Performance with Different Numbers of USVs and UAVs}
Fig. \ref{f4}(a) shows the average execution delay as the number of USVs increases. GAI-HAPPO maintains low delays, demonstrating its scalability and efficiency. Transformer-HAPPO performs similarly but slightly lags at higher USV counts. GAN-HAPPO shows moderate scalability, while HAPPO and HAA2C exhibit significantly higher delays, struggling to handle increased loads.

Fig. \ref{f4}(b) presents the average reward as the number of USVs increases. GAI-HAPPO consistently achieves the highest rewards, showcasing its ability to optimize decision-making and resource allocation. Transformer-HAPPO performs comparably but degrades slightly with larger USV numbers. GAN-HAPPO provides moderate rewards, while HAPPO and HAA2C show poor adaptability to complex environments.

Fig. \ref{f4}(c) depicts the average execution delay for increasing numbers of UAVs. GAI-HAPPO achieves the shortest delays consistently, demonstrating efficient utilization of additional UAV resources. Transformer-HAPPO performs well but lags slightly behind GAI-HAPPO. GAN-HAPPO shows moderate scalability, while HAPPO and HAA2C experience significantly higher delays, with HAA2C performing the worst.

Fig. \ref{f4}(d) highlights the average rewards as the number of UAVs increases. GAI-HAPPO achieves the highest rewards across all scenarios, reflecting its superior resource allocation and system performance. Transformer-HAPPO follows closely with competitive results. GAN-HAPPO shows moderate performance, while HAPPO and HAA2C struggle with resource utilization and task execution, especially as UAV numbers increase.

\section{Conclusions}\label{s5}
In this paper, we design a cooperative UAV-GS-based MEC model to meet the computational and coverage demands of USVs in complex applications, such as environmental monitoring and maritime search and rescue.
The framework efficiently supports USVs in completing computational tasks by combining the mobility and flexibility of UAVs with the robust computational power of GSs.
To address the challenges of task uncertainties, USV trajectory unpredictability, device heterogeneity, and limited computational resources, we formulate a joint optimization problem involving task offloading and UAV trajectory planning, aimed at minimizing total execution time. 
In addition, we develop the GAI-HAPPO algorithm. GAI-HAPPO incorporates generative AI models to improve the actor network’s ability to model complex environments, extract high-level features, and adapt to dynamic conditions, while also stabilizing the critic network to mitigate MARL instability.
Extensive simulations show that our approach outperforms existing baseline methods by 22.8\%, demonstrating its superior effectiveness in optimizing collaboration within MEC scenarios.

\bibliography{references}
\end{document}